\documentclass[letterpaper]{article}

\usepackage[preprint]{aaai2027}
\usepackage[hyphens]{url}
\usepackage{graphicx}
\usepackage{natbib}
\usepackage{microtype}
\usepackage{amsmath,amssymb,amsfonts}
\usepackage{booktabs}
\usepackage{xcolor}
\usepackage{colortbl}
\usepackage{tikz}
\usetikzlibrary{decorations.pathreplacing,positioning,calc}

\newcommand{\dz}{\Delta\mathbf{z}}
\newcommand{\zz}{\mathbf{z}}
\newcommand{\Scal}{\mathcal{S}}
\newcommand{\Ocal}{\mathcal{O}}
\newcommand{\enc}{\Phi}
\newcommand{\reals}{\mathbb{R}}
\newcommand{\softmax}{\mathrm{softmax}}
\newcommand{\softplus}{\mathrm{softplus}}

\newcommand{\PaperAuthors}{%
  Zhongyao Wang\textsuperscript{1},
  Wanli Ouyang\textsuperscript{2,*},
  Taoyong Cui\textsuperscript{2,3,*},
  Pheng Ann Heng\textsuperscript{3}%
}
\newcommand{\PaperAffiliations}{%
  \textsuperscript{1}Fudan University\\
  \textsuperscript{2}MMLab, The Chinese University of Hong Kong\\
  \textsuperscript{3}Department of Computer Science and Engineering,
  The Chinese University of Hong Kong\\
  \textsuperscript{*}Corresponding authors:
  \texttt{wlouyang@ie.cuhk.edu.hk, cty21@tsinghua.org.cn}%
}
\newcommand{\PaperPdfAuthors}{%
  Zhongyao Wang; Wanli Ouyang; Taoyong Cui; Pheng Ann Heng%
}

\title{Support$\times$Operation Factorization:\\
Compositional Readout of Frozen Vision Encoders\\
under Controlled Interventions}

\author{\PaperAuthors}
\affiliations{\PaperAffiliations}

\begin{document}
\maketitle

\begin{abstract}
Compositional analysis of frozen vision encoders should determine both
\emph{what} changed and \emph{where} it changed.  Standard factor probes
score these axes separately, however, and can reward multiple
operations that reuse the same predicted slot.  We call this failure
\emph{operation laundering}.  We introduce an injectively aligned
leave-one-cell-out protocol over support$\times$operation grids and
SO-OPF, a readout that factors cell energy into support salience and a
competitive operation posterior.  This formulation separates two
questions that aggregate scores conflate: whether the carrier composes
held-out bindings when the grid is known, and whether that grid can be
recovered from flat cell labels.  With frozen DINOv3 features, known
factorial assignment reaches $0.874$ injective accuracy on
Shapes3D-Extended and $0.799$ on globally image-disjoint COCO; learning
the assignment from flat labels reaches $0.769$ and $0.762$,
respectively.  Under matched axis-aware supervision on Shapes3D, the
factored carrier improves learned-assignment accuracy from $0.653$ to
$0.841$ over a dense carrier and eliminates its laundering gap.
SigLIP2 replicates the COCO separation.  A rebuilt MuJoCo substrate
exposes a boundary: learned-assignment accuracy is $0.569$ with DINOv3
and $0.484$ with SigLIP2, with substantial slot collapse.  Thus
factored readout and injective evaluation recover held-out bindings on
two substrates while exposing, rather than hiding, a renderer-specific
failure boundary; they do not establish universal recovery from flat
labels.
\end{abstract}

\section{Introduction}
\label{sec:intro}

Self-supervised visual encoders such as
DINOv3~\citep{oquab2025dinov3}, I-JEPA~\citep{assran2023self}, and
V-JEPA~\citep{bardes2024revisiting} record systematic image edits in
their patch-token features.  Many analyses treat these traces as
\emph{factors of variation}: global directions in latent space, as in
principal components or linear concept probes.  Disentanglement methods
\citep{higgins2017betavae,kim2018disentangling} adopt this view
explicitly, while predictive world-model approaches
\citep{lecun2022path,garrido2024learning} motivate analyses of how
controlled changes are encoded in latent space.

A global direction does not explicitly bind a change class to the
image region where that change occurred.  A readout that cannot
distinguish floor-hue from wall-pattern edits---or that credits both
through the same representational slot---fails silently under
single-factor metrics.  In a world model that must independently
manipulate scene attributes, such confusion means that controlling one
factor inadvertently alters another.  Held-out-combination tests
can reveal this limitation where single-factor evaluations
miss it~\citep{montero2021role,schott2022visual}.

\begin{figure*}[t]
\centering
\includegraphics[width=\textwidth]{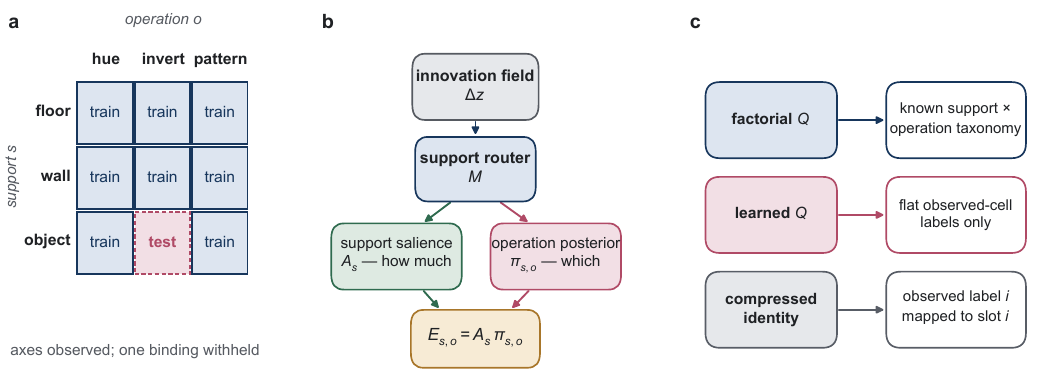}
\caption{\textbf{Evaluation, carrier, and assignment are separate
design choices.}
\textbf{a}, Leave-one-cell-out evaluation withholds one binding while
retaining its support and operation in other cells.
\textbf{b}, SO-OPF factors support salience from a competitive
operation posterior.
\textbf{c}, The structural input each assignment arm receives.  Only
the learned arm tests structure recovery without the axis taxonomy;
compressed identity is a control that maps label order onto slots
directly, and we show below that this order does not supply a
factorial grid.}
\label{fig:teaser}
\end{figure*}

We ask a readout-design question: given examples such as floor-hue
and wall-pattern edits, can a structured readout reuse \emph{where}
and \emph{what} axes to route a support--operation pairing absent from
training?  To answer it, we define factors as cells in a
support$\times$operation grid and evaluate under a
leave-one-cell-out protocol (Figure~\ref{fig:teaser}).  The
\emph{innovation field}
$\dz=\zz_{\mathrm{int}}-\zz_{\mathrm{src}}$ records how a frozen
encoder's patch-token map changes under an intervention; a cell binds
a spatial support~$s$ to an operation~$o$.

The metric must also reject degenerate slot sharing.  A many-to-one
post-hoc alignment can credit two operations through the same slot---a
failure we call \emph{operation laundering}.  We instead use a
Hungarian injective assignment~\citep{kuhn1955hungarian}, estimated
only from training-pair energies, so every ground-truth axis value
must claim a distinct slot.

SO-OPF separates support salience from operation identity:
$E_{s,o}=A_s\pi_{s,o}$.  Here $A_s$ measures how much innovation falls
under support~$s$, while $\pi_{s,o}$ is a competitive posterior over
operations computed from scale-normalized innovation features.  This
factored carrier makes its structural assumption explicit.  It does
not, by itself, specify how flat observed-cell labels correspond to a
factorial grid.

The separation of carrier from assignment motivates two experimental
questions.  In every experiment, the grid dimensions and flat
observed-cell labels are known.  First, \emph{carrier capacity}: if the
support--operation taxonomy is also supplied, can the factored readout
compose held-out cells?  Second, \emph{assignment recovery}: can the
mapping from flat labels to grid locations be estimated without
separate axis labels?  The paired comparison of these two conditions
across datasets reveals where assignment estimation costs accuracy and
where it does not.

Two additional controls locate the boundaries of the claim.  A
\emph{compressed-label assignment}---observed label~$i$ maps to
slot~$i$---tests whether label order alone can supply a factorial
grid.  A rebuilt \emph{MuJoCo substrate} tests whether compositional
routing transfers across renderers.  Together, the assignment control
and the substrate boundary delimit what SO-OPF can and cannot recover.

Our contributions are:
\begin{enumerate}
\itemsep0pt
\item[{\bf C1}] a support$\times$operation definition and a
  leave-one-cell-out metric with injective binding that exposes
  operation laundering;
\item[{\bf C2}] the SO-OPF carrier, which separates support salience
  from operation identity and outperforms a dense carrier under
  matched axis-aware supervision;
\item[{\bf C3}] an empirical separation of known factorial assignment,
  assignment learned from flat cell labels, and compressed label order
  across Shapes3D, MuJoCo, and globally image-disjoint COCO, with a
  second-encoder replication on COCO and MuJoCo.
\end{enumerate}

\section{Related Work}
\label{sec:related}

\paragraph{Disentangled representation learning.}
$\beta$-VAE \citep{higgins2017betavae}, FactorVAE
\citep{kim2018disentangling}, and their descendants encourage factors
to align with statistically independent latent coordinates.
\citet{locatello2019challenging} show that unsupervised
disentanglement is underdetermined without inductive bias; subsequent
work reveals that even well-disentangled representations fail
compositional generalization
\citep{montero2021role,schott2022visual}, and that factor correlations
in the training data further entangle what is learned
\citep{trauble2021disentangled}.  \citet{higgins2018towards} ground
disentanglement in symmetry transformations, closer to our
operation axis.  One source of
the failure may be the global-direction definition itself, which
provides no explicit binding between a spatial region and a
transform class.

\paragraph{JEPA and predictive world models.}
Joint-Embedding Predictive Architectures \citep{lecun2022path} learn
latent predictions of future or masked content.  I-JEPA
\citep{assran2023self}, V-JEPA \citep{bardes2024revisiting}, and
Image World Models \citep{garrido2024learning} demonstrate strong
latent spaces but do not primarily study whether spatial factors
compose under held-out bindings.
Recent work decomposes JEPA latent spaces into subspaces or
sparse components for improved structure
\citep{chen2025subspace,agrawal2025rectified}, but does not evaluate
compositional generalization of spatial factors.  We use frozen DINOv3
and SigLIP2 vision transformers as substrates and ask what readout makes
their latent innovation fields compositionally recoverable under a
specified intervention and supervision protocol.

\paragraph{Object-centric and slot models.}
MONet \citep{burgess2019monet}, IODINE \citep{greff2019multi},
Slot Attention \citep{locatello2020object}, and its video extensions
\citep{kipf2022conditional,singh2022illiterate} decompose scenes into
object slots, capturing the \emph{where} axis of our framework.
A slot carries the object together with whatever happens to
it; we model the operation as its own separately routed axis.
Tensor-product representations
\citep{smolensky1990tensor} bind roles to fillers via outer products;
our support$\times$operation factorization is analogous, with spatial
masks as role selectors and operation posteriors as fillers.

\paragraph{Probing and concept analysis.}
Linear probes \citep{alain2017understanding,belinkov2022probing},
concept activation vectors \citep{kim2018interpretability}, and
control tasks \citep{hewitt2019control} read
out properties from frozen representations but typically test one factor
at a time; Network Dissection \citep{bau2017network} localizes
concepts to units.  Our readout is a
structured probe that reads out
\emph{interventions} (the innovation field) under a
\emph{compositional} protocol, and is scored on a held-out
support$\times$operation binding rather than on single-attribute
classification.

\paragraph{Compositional generalization.}
Systematic generalization tests
\citep{lake2018generalization,keysers2020measuring} evaluate whether a
model handles unseen compositions of known primitives.
\citet{wiedemer2024compositional} provide theoretical conditions for
compositional generalization from first principles.
Compositional zero-shot learning (CZSL) evaluates unseen
attribute--object pairs \citep{misra2017red}; our
support$\times$operation grid is structurally analogous, with the
difference that we evaluate a readout atop a frozen encoder
rather than an end-to-end classifier.  Our
leave-one-cell-out protocol instantiates systematic generalization
for spatial-visual factors.

\paragraph{Causal representation learning.}
Causal representation learning
\citep{scholkopf2021toward,vonkugelgen2021self} and nonlinear ICA
\citep{hyvarinen2019nonlinear} study whether latent variables can be
identified from interventions or auxiliary signals; their evaluation
criterion is identifiability of individual coordinates.  We address a
complementary question: given a frozen encoder with known intervention
structure, can a readout \emph{compose} spatial bindings that were
absent from training?  Our setting is structurally simpler (no
identifiability proof is attempted), but adds a compositional
held-out-binding evaluation that identifiability analyses do not
require.

\section{Support$\times$Operation Factorization}
\label{sec:definition}

\paragraph{Frozen encoder and token maps.}
Let $\enc$ be a frozen vision transformer that maps an image to a grid
of patch tokens $\zz \in \reals^{N \times d}$.  We use DINOv3-ViT-L
\citep{oquab2025dinov3} for the main experiments and SigLIP2-L/16
\citep{tschannen2025siglip2} for the cross-encoder replication.  Both
produce 1,024-dimensional tokens; their respective token grids contain
196 and 256 patches.  The encoder is never trained or fine-tuned; all
learning occurs in the readout.

\paragraph{The innovation field.}
Given a source image $x_{\mathrm{src}}$ and an intervened image
$x_{\mathrm{int}}$ that differs by a single visual edit (e.g., a
shifted floor hue), the \emph{innovation field} is
\begin{equation}
  \dz = \enc(x_{\mathrm{int}}) - \enc(x_{\mathrm{src}})
  \;\in\; \reals^{N \times d}.
  \label{eq:innovation}
\end{equation}
It records how the frozen token map changes under one intervention and
serves as the sole input to the readout.

\paragraph{Supports and operations.}
We decompose the innovation field along two axes:
\begin{itemize}
\itemsep0pt
\item A \textbf{support} $s \in \Scal$ is a ground-truth spatial
  region---a subset of patches where $\dz$ typically concentrates:
  \emph{where} the change occurs (floor, wall, or object).
\item An \textbf{operation} $o \in \Ocal$ is a transform class:
  \emph{what} the change is (hue rotation, color inversion, or stripe
  overlay).
\end{itemize}

\paragraph{Predictive factor = support$\times$operation cell.}
A predictive factor is a \emph{cell} $(s, o)$ in the factorial grid
$\Scal \times \Ocal$.  With $|\Scal|{=}3$ supports and $|\Ocal|{=}3$
operations, the grid contains nine cells.  This definition departs from
analyses that equate a factor with a single global direction in latent
space: it binds a spatial locus to a transform identity.  The
evaluation tests cell routing; mask overlap with the ground-truth
region is reported separately and is not required by the cell-accuracy
metric.

\paragraph{Contrast with global-direction factors.}
Under the global-direction view, each factor is a vector
$\mathbf{v}_k \in \reals^d$ with $\langle \zz, \mathbf{v}_k
\rangle$ tracking the $k$-th ground-truth attribute.
$\mathbf{v}_k$ is shared across patches and does not itself encode a
support--operation binding, so operations on \emph{different}
supports (floor-hue and wall-pattern) may be difficult to route
independently.  The cell view gives every support a mask and every
operation a slot, making local routing directly testable.

\section{Compositional Metric with Injective Binding}
\label{sec:metric}

The evaluation asks whether a readout routes an unseen
support$\times$operation binding while detecting solutions that share
slots between operations.

\paragraph{Leave-one-cell-out protocol.}
For each cell $(s^*,o^*)$ in an $|\Scal|\times|\Ocal|$ grid, we train
on the other $|\Scal||\Ocal|-1$ cells.  The support $s^*$ remains
paired with other operations and $o^*$ remains paired with other
supports; only their binding is held out.  Within each observed cell,
a seeded permutation allocates 64\% of pairs to optimization, 16\% to
restart validation, and leaves 20\% unused.  Every pair from the
held-out cell is test-only.

\paragraph{Joint prediction and axis alignment.}
For a batch of test pairs, the readout produces
$E\in\reals^{B\times|\Scal|\times|\Ocal|}$ and predicts
\begin{equation}
  (\hat{s},\hat{o})=\arg\max_{s,o}E_{b,s,o}.
  \label{eq:joint_argmax}
\end{equation}
The internal energy axes may be permuted, especially when the
cell-to-grid assignment is learned.  From optimization pairs only, we
therefore compute mean support- and operation-marginal energy matrices
$C^{\mathrm{sup}}$ and $C^{\mathrm{op}}$ and bind each
ground-truth axis value to an energy slot.  This alignment is used for
evaluation, never for training or restart selection.

\paragraph{Operation laundering.}
A non-injective alignment maps every axis value independently to its
highest-energy slot.  Two operations can then share a slot and both
receive credit.  We call this failure \emph{operation laundering}.
Injective alignment prevents it by requiring each ground-truth axis
value to claim a distinct slot.

\paragraph{Hungarian injective assignment.}
The injective cell accuracy (INJ) constrains each axis map to a
bijection:
\begin{equation}
 \sigma^*_{\mathrm{ax}}
 =\arg\max_{\sigma}\sum_v C^{\mathrm{ax}}_{v,\sigma(v)},
 \qquad \mathrm{ax}\in\{\mathrm{sup},\mathrm{op}\}.
\end{equation}
The Hungarian algorithm~\citep{kuhn1955hungarian} solves this
assignment.  A held-out example is correct when its joint argmax is
$(\sigma^*_{\mathrm{sup}}(s^*),
\sigma^*_{\mathrm{op}}(o^*))$.  INJ is averaged first over all nine
held-out cells within a seed and then across seeds.

\begin{figure}[t]
\small
\centering
\fbox{\parbox{0.93\columnwidth}{%
\textbf{Algorithm 1}\; Leave-one-cell-out injective evaluation
\vspace{2pt}\hrule\vspace{4pt}
\textbf{Input:} cell-partitioned data $\mathcal D$, seeds, restarts,
assignment mode $Q$\\
\textbf{Output:} per-cell and mean INJ

\vspace{3pt}
\textbf{for} seed $k$ and held-out cell $h=(s^*,o^*)$ \textbf{do}\\
\quad Split every observed cell into 64\% optimization,
16\% validation, 20\% unused\\
\quad Reserve all pairs in $h$ for test\\
\quad \textbf{for} restart $r$ \textbf{do}\\
\qquad Fit readout and $Q$ when $Q$ is learned\\
\qquad Score flat observed-cell accuracy on validation pairs\\
\quad \textbf{end for}\\
\quad Select the highest-validation restart \hfill
\textrm{\scriptsize// no Hungarian/test labels}\\
\quad From optimization pairs, compute
$C^{\mathrm{sup}},C^{\mathrm{op}}$\\
\quad Compute $\sigma^*_{\mathrm{sup}},\sigma^*_{\mathrm{op}}$
with Hungarian assignment\\
\quad Evaluate the joint argmax on held-out-cell pairs\\
\textbf{end for}\\
\textbf{return} seed- and cell-aggregated INJ
}}
\label{alg:metric}
\end{figure}

\paragraph{Restart selection.}
Every restart is selected by flat observed-cell validation accuracy
under the assignment fitted in that restart.  Factorial assignment is
fixed; learned assignment is optimized jointly with the readout.
Validation does not use Hungarian alignment, held-out pairs, or
held-out labels.  The supervised MLP, attention, and oracle-operation
baselines use native joint validation accuracy.

\paragraph{Diagnostics.}
One fit is a (seed, held-out cell) combination, giving 90 fits in a
10-seed $3\times3$ experiment.  The \emph{collapse rate} is the
fraction of fits whose held-out-cell accuracy is below a predeclared
threshold of 0.5; we also report thresholds 0.4 and 0.6 in
the supplementary material.  Assignment recovery compares a learned
$Q$ with the true factorial grid up to global permutations of the two
axes.  It is a diagnostic of recovered grid structure, not an input to
training or scoring.

\section{Method: Factored Energy and Grid Assignment}
\label{sec:method}

\subsection{The SO-OPF Readout}
\label{sec:soopf}

A readout must separate how much change falls under each support
from which operation produced it.  A dense carrier that pools
per-channel squared innovation couples these two roles, letting
different operations share response channels.  SO-OPF factors them
apart, mapping an innovation field
$\dz\in\reals^{N\times d}$ to a
support$\times$operation energy grid
$E\in\reals^{|\Scal|\times|\Ocal|}$.  A learned router first assigns
patches to support slots:
\begin{equation}
 M_{s,n}=\softmax_s\!\left(
 \left\langle W_z\zz_{\mathrm{src},n}+\mathbf p_n,\mathbf q_s\right\rangle
 \right),
 \label{eq:mask}
\end{equation}
where $W_z$, positional embeddings $\mathbf p_n$, and support queries
$\mathbf q_s$ are learned.  The router receives no pixel mask.

The readout then separates the amount of change from its identity:
\begin{equation}
 E_{s,o}=A_s\,\pi_{s,o},
 \label{eq:factorop}
\end{equation}
with
\begin{align}
 A_s &= \sum_n M_{s,n}\,\mathrm{sal}_n,\\
 \pi_{s,o} &=
 \softmax_o\!\left(f_\theta(\mathbf h_s)/\tau\right),\\
 \mathbf h_s &=
 \frac{\sum_n M_{s,n}\mathrm{sal}_n\,\mathrm{LN}(\dz_n)}
 {\sum_n M_{s,n}\mathrm{sal}_n},
\end{align}
where $\mathrm{sal}_n=\|\dz_n\|_2^2$.  Thus $A_s$ retains
support-level innovation mass, while $\pi_{s,o}$ competes over
operation identities using scale-normalized features.
$f_\theta$ is a one-hidden-layer MLP with 64 GELU units, and
$\tau$ is annealed from 2.0 to 0.5.

\subsection{Assignment Semantics}
\label{sec:assignment}

For a held-out cell, let
$K=|\Scal||\Ocal|-1$ be the number of observed cells.  Each training
pair has a flat label $y\in\{0,\ldots,K-1\}$.  A row-stochastic
assignment
$Q\in[0,1]^{K\times|\Scal||\Ocal|}$ maps those labels to energy-grid
locations.  With normalized energy
$\mathbf{sh}=E/\sum_{s,o}E_{s,o}$, the observed-cell loss is
\begin{equation}
 \mathcal L_{\mathrm{cell}}
 =\mathrm{CE}\!\left(10\,\mathbf{sh}Q^\top,y\right).
 \label{eq:cell_loss}
\end{equation}
The factor 10 prevents energy-share logits from remaining nearly
uniform.

We distinguish three assignment semantics.
\begin{itemize}
\itemsep0pt
\item \textbf{Factorial $Q$.}  Row $y$ is a one-hot vector at the
  true grid location $(s_y,o_y)$.  This arm receives the known
  support--operation taxonomy and is an upper reference.
\item \textbf{Learned $Q$.}  Each row is $\softmax(\mathbf a_y)$ and
  is optimized jointly with the readout.  Grid dimensions are known,
  but the core objective does not receive separate support or
  operation labels.
\item \textbf{Compressed identity.}  Observed label $y$ maps to grid
  column $y$, leaving the final column unused.  Because the held-out
  factorial cell changes across folds, this convention generally
  preserves neither grid axis.  It tests the literal label-order
  interpretation, not a reindexing of factorial $Q$.
\end{itemize}

Held-out cells retain their true factorial coordinates under factorial
$Q$; only compressed identity reserves the final slot.
The metric's Hungarian alignment subsequently evaluates energy-slot
semantics from optimization pairs.  It does not repair the training
assignment or participate in restart selection.

\subsection{Core and Axis-Aware Objectives}
\label{sec:training}

The core protocol uses
\begin{equation}
\begin{aligned}
 \mathcal L_{\mathrm{core}}
 &=\mathcal L_{\mathrm{cell}}
 +\lambda_{\mathrm{mask}}\mathcal L_{\mathrm{mask}},\\
 \lambda_{\mathrm{mask}}&=1,
\end{aligned}
\label{eq:core_objective}
\end{equation}
where $\mathcal L_{\mathrm{mask}}$ is the negative entropy of mean
support usage and discourages an empty support slot.  Localization,
assignment-balance, axis-balance, and operation-injectivity weights
are zero.  The core learned-$Q$ arm therefore uses flat cell labels
without native axis labels.

Some carrier comparisons use a separate axis-aware enhanced protocol:
\begin{equation}
\begin{split}
 \mathcal L_{\mathrm{enh}}={}&
 \mathcal L_{\mathrm{cell}}+\mathcal L_{\mathrm{loc}}
 +0.1\,\mathcal L_{\mathrm{bal}}\\
 &+0.5\,\mathcal L_{\mathrm{axis}}
 +\mathcal L_{\mathrm{mask}}+\mathcal L_{\mathrm{opinj}} .
\end{split}
\label{eq:enh_objective}
\end{equation}
$\mathcal L_{\mathrm{loc}}$ penalizes energy outside the cell assigned
to a flat label and requires assigned mass at least $\rho=0.6$.
$\mathcal L_{\mathrm{bal}}$ and $\mathcal L_{\mathrm{axis}}$
discourage unused assignment columns and axes.
$\mathcal L_{\mathrm{opinj}}$ forms mean operation-slot profiles from
\emph{native operation labels} and penalizes their off-diagonal cosine
similarity.  Results using this loss compare carrier design under
axis-aware supervision; they are not evidence for discovery from flat
labels alone.

\subsection{Dense Carrier Baseline}
\label{sec:carrier}

For comparison, a dense nonnegative carrier pools squared innovation
channels independently for every operation:
\begin{equation}
 E_{s,o}^{\mathrm{dense}}
 =\sum_n M_{s,n}\sum_c
 \softplus(W_{o,c})(\dz_{n,c})^2 .
 \label{eq:dense}
\end{equation}
It couples operation identity to change magnitude.  The carrier
comparison in the experiments tests this form against
Eq.~\ref{eq:factorop} under matched axis-aware supervision.  The core
learned-$Q$ experiment separately tests assignment recovery from flat
labels.

\section{Experiments}
\label{sec:experiments}

\subsection{Substrates and Protocol}
\label{sec:substrates}

The three substrates separate controlled factor composition, renderer
sensitivity, and behavior on natural images.

\paragraph{Shapes3D-Extended.}
We extend Shapes3D~\citep{kim2019shapes3d} with three supports
(floor, wall, object) and three operations (hue, invert, pattern),
giving nine cells with 350 source--intervention pairs per cell.
Hue selects a different native hue uniformly; inversion uses
$255-x$ within the support; pattern adds randomly oriented stripes
with frequency in $[0.15,0.45)$ and integer amplitude 40--89.

\paragraph{MuJoCo (renderer-sensitivity test).}
To test whether compositional routing transfers beyond a single
renderer, we build a second substrate using
MuJoCo~\citep{todorov2012mujoco} with box geometry and separate camera
and lighting settings, with 350 pairs per cell.  Hue rotates the
target material by a value sampled uniformly from $[0.25,0.75)$ turns,
and inversion changes the material before rerendering.  Pattern
remains a masked pixel-space stripe overlay with frequency
$[0.08,0.22)$ and amplitude 45--94.

\paragraph{Image-disjoint COCO.}
We edit person, vehicle, and background regions in
COCO~\citep{lin2014coco}.  Each cell contains 300 pairs, and all 2,700
source images are globally unique across cells; consequently no source
used by an observed cell appears in the held-out cell.  Hue is sampled
uniformly from $[0.15,0.85)$ turns, with the same inversion and stripe
families as Shapes3D.

\paragraph{Features and optimization.}
The headline experiments use frozen DINOv3-ViT-L/16 features
\citep{oquab2025dinov3} from
\path{dinov3-vitl16-pretrain-lvd1689m}; the cross-encoder replication
uses frozen SigLIP2-L/16-256 features
\citep{tschannen2025siglip2} from
\path{siglip2-large-patch16-256}.  We use each checkpoint's bundled
image processor and patch tokens, giving $(N,d)=(196,1024)$ for DINOv3
and $(256,1024)$ for SigLIP2.  Readouts use Adam with learning rate
$5\times10^{-3}$ for 2,000 steps and five restarts.  The main
experiments use seeds 0--9; data construction uses seed 0 for COCO
and MuJoCo.  DINOv3 and SigLIP2 runs were executed on NVIDIA H200 and
H100 GPUs, respectively.
Restart selection and the 64\%/16\%/20\% observed-cell split follow
the metric protocol above.

\paragraph{Reporting.}
We aggregate the nine held-out-cell accuracies within each seed and
report the mean and population standard deviation across ten seeds.
All $\pm$ values below are therefore across-seed SD.  Paired tests use
the ten aligned seed-level means.

\paragraph{Comparators.}
We compare a bilinear structured carrier
\citep{tenenbaum2000separating} using factorial assignment,
two-layer MLP and attention probes on mean-pooled or token-level
$\dz$, and an oracle-region operation classifier.  The supervised
probes receive native axis labels.  Randomly initialized frozen ViT
and raw-pixel carriers are additionally evaluated on COCO.

\begin{figure}[t]
  \centering
  \includegraphics[width=0.88\columnwidth]{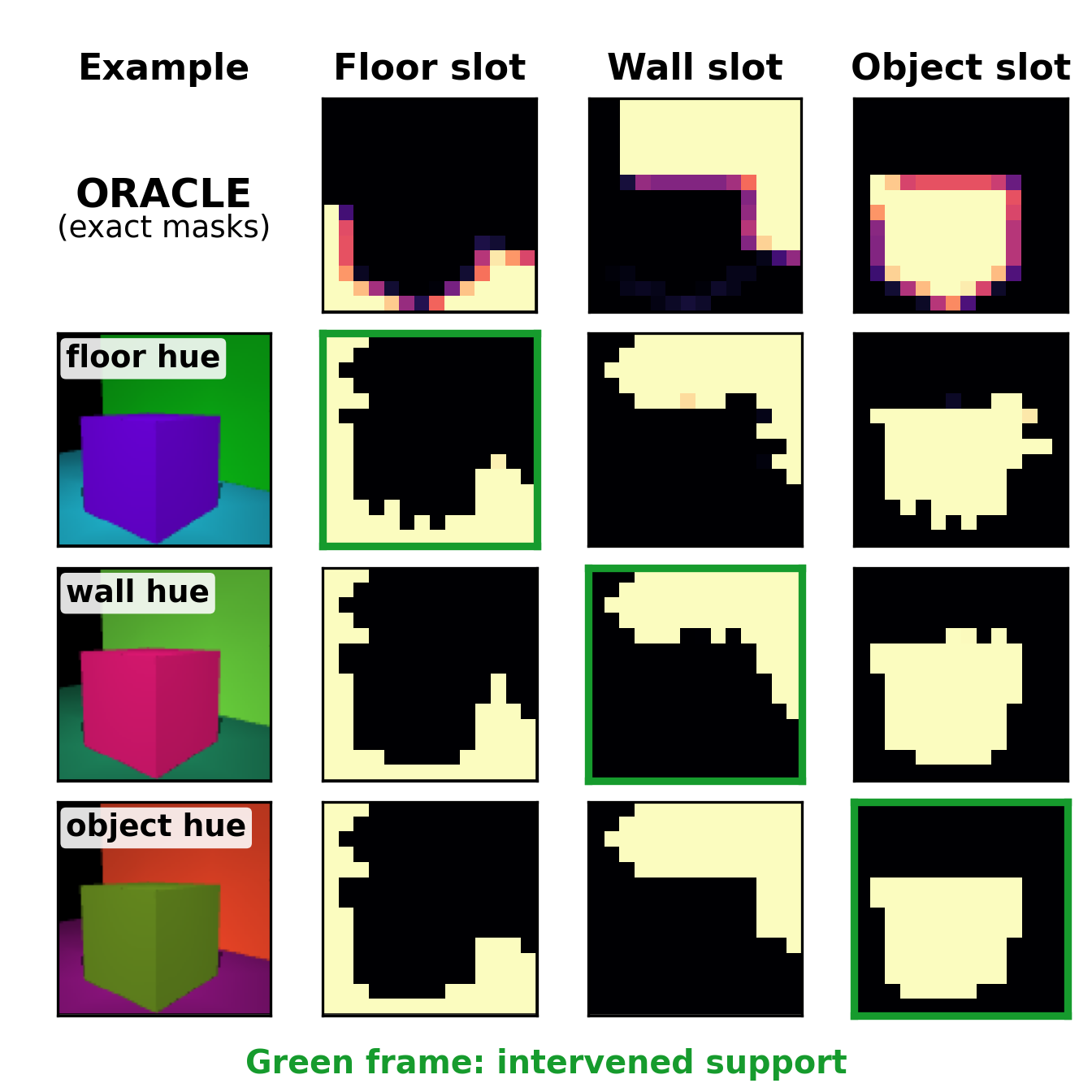}
  \caption{\textbf{MuJoCo support-router diagnostic.}
  Exact support masks are shown above three hue-intervention examples;
  the remaining panels show the learned support slots.  Green frames
  mark the intervened support.  After matching slots to supports, mean
  mask IoU is 0.821 and assignment agreement is 0.903 over 630 held-out
  examples.  This panel diagnoses support localization; it is not a
  learned-$Q$ factorial-assignment result.}
  \label{fig:maskviz}
\end{figure}

\subsection{Known Structure and Learned Structure}
\label{sec:headline}

\begin{figure*}[t]
\centering
\includegraphics[width=\textwidth]{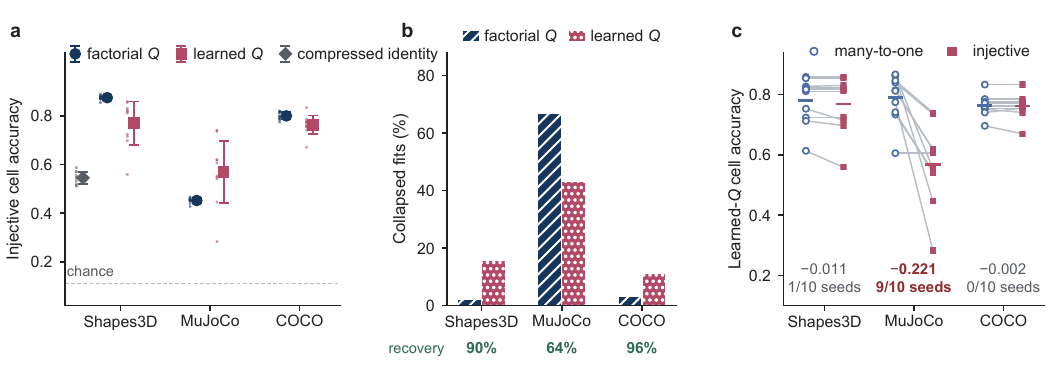}
\caption{\textbf{DINOv3 assignment comparison.}
\textbf{a}, Injective cell accuracy for factorial, learned, and
compressed-identity assignments.  Each seed is first averaged over
nine held-out cells; small dots are the ten seed means, and markers
with bars are their mean and population SD.  Compressed identity is a
Shapes3D-only label-order control.  The dashed rule marks chance
($1/9$).
\textbf{b}, Collapse rate over 90 seed--cell fits; the green row
beneath the axis reports learned-$Q$ assignment recovery.
\textbf{c}, Many-to-one and injective learned-$Q$ scores for each of
the ten seeds, joined in pairs; short horizontal rules are the arm
means.  MuJoCo loses accuracy under injective scoring in 9 of 10
seeds, whereas Shapes3D and COCO are essentially unaffected---the
laundering the aggregate many-to-one score conceals is a per-seed
effect, not an artifact of averaging.
Factorial $Q$ receives the support--operation taxonomy, whereas
learned $Q$ uses flat observed-cell labels only.}
\label{fig:assignment_audit}
\end{figure*}

\begin{table}[!htbp]
\centering
\caption{\textbf{Assignment semantics under the core protocol.}
INJ is injective cell accuracy; Min is the minimum held-out-cell mean;
Coll. is the fraction of 90 seed--cell fits below 0.5; Rec. is
factorial assignment recovery up to axis permutations.  Values after
$\pm$ are across-seed SD ($n=10$).  Figure~\ref{fig:assignment_audit}
shows the DINOv3 rows.  Factorial $Q$ receives the axis taxonomy;
learned $Q$ uses flat cell labels only.}
\label{tab:headline}
\small
\setlength{\tabcolsep}{3.4pt}
\begin{tabular}{@{}lcccc@{}}
\toprule
\rowcolor{black!6}
Assignment & INJ & Min & Coll.$\downarrow$ & Rec.$\uparrow$\\
\midrule
\multicolumn{5}{@{}l}{\textit{Shapes3D, DINOv3}}\\
factorial & .874$\pm$.009 & .613 & .022 & n/a\\
learned & .769$\pm$.090 & .560 & .156 & .900\\
compressed identity & .545$\pm$.025 & .016 & .444 & n/a\\
\midrule
\multicolumn{5}{@{}l}{\textit{MuJoCo, DINOv3}}\\
factorial & .452$\pm$.011 & .090 & .667 & n/a\\
learned & .569$\pm$.127 & .124 & .433 & .643\\
\multicolumn{5}{@{}l}{\textit{MuJoCo, SigLIP2}}\\
factorial & .476$\pm$.008 & .045 & .656 & n/a\\
learned & .484$\pm$.167 & .003 & .511 & .676\\
\midrule
\multicolumn{5}{@{}l}{\textit{COCO, DINOv3}}\\
factorial & .799$\pm$.015 & .524 & .033 & n/a\\
learned & .762$\pm$.039 & .533 & .111 & .961\\
\multicolumn{5}{@{}l}{\textit{COCO, SigLIP2}}\\
factorial & .826$\pm$.012 & .563 & .011 & n/a\\
learned & .751$\pm$.074 & .448 & .078 & .965\\
\bottomrule
\end{tabular}
\end{table}

Factorial assignment establishes what the SO-OPF carrier can do when
the grid taxonomy is supplied.  It composes held-out cells on
Shapes3D and COCO.  Learned assignment asks the
stronger flat-label question.  On Shapes3D, its raw and injective
scores are 0.780 and 0.769; the paired factorial-minus-learned raw
difference is 0.095 ($t=4.01$, two-sided $p=0.0031$).  On COCO, its
raw and injective scores are 0.765 and 0.762.  Recovery rates of 90.0\%
and 96.1\% show that most fits discover an equivalent grid, but the
score gaps and across-seed variation rule out lossless equivalence to
known structure.

The COCO conclusion replicates with SigLIP2.  Factorial and learned
assignments reach 0.826 and 0.751 INJ; their paired difference is 0.075
($t=3.33$, two-sided $p=0.0087$).  Learned assignment recovers an
equivalent grid in 96.5\% of fits.  Thus changing the frozen encoder
preserves both the strong carrier result and the measurable cost of
estimating its assignment from flat labels.

\paragraph{Injective scoring exposes operation laundering.}
Without injective binding, a many-to-one alignment can credit two
operations through the same slot and inflate accuracy.  This effect is
most visible on MuJoCo: learned assignment scores 0.790 under
many-to-one alignment but drops to 0.569 after injective scoring---a
gap of 0.221 that the aggregate many-to-one number conceals entirely
(Figure~\ref{fig:assignment_audit}c).  The paired seed-level view
shows this is not an averaging artifact: 9 of 10 MuJoCo seeds lose
more than 0.05 accuracy under injective scoring (paired $t=4.71$),
against 1 of 10 on Shapes3D and 0 of 10 on COCO.
SigLIP2 exhibits the same
pattern: its learned score falls from 0.629 to 0.484.  These gaps
confirm that reporting only many-to-one accuracy can hide severe slot
collapse.

\paragraph{Label-order control.}
Compressed identity tests whether label order alone supplies a grid.
Mapping the $i$-th observed flat label to the $i$-th slot scores 0.545
and collapses in 44.4\% of fits.  Because a different cell is held out
in every fold, the compressed mapping assigns different labels to the
same row or column across folds; this inconsistency means that label
order is not a harmless reindexing of the factorial grid but an
assignment that generally violates its axis structure.

\paragraph{Renderer boundary.}
The rebuilt MuJoCo substrate changes the cross-renderer conclusion.
With DINOv3, factorial assignment reaches only 0.452; learned
assignment rises to 0.790 before injective alignment but falls to
0.569 after it, with 43.3\% collapse.  SigLIP2 does not repair the
boundary: factorial and learned INJ are 0.476 and 0.484
($p=0.877$), and the learned score drops from 0.629 to 0.484 under
injective alignment.  Its 51.1\% collapse and minimum cell mean of
0.003 show severe fold dependence.  Per-cell breakdowns reveal that
pattern cells remain near-perfect ($>0.99$) while hue and inversion
cells drive the failure---consistent with material-space operations
producing less discriminable token-level signatures than pixel-space
overlays.  The boundary is stable across collapse thresholds 0.4--0.6
(SI~\S3).  MuJoCo therefore bounds, rather than confirms, an encoder-
or renderer-independent claim.

\subsection{Baselines and Carrier Comparison}
\label{sec:carrier_exp}

\begin{table}[!htbp]
\centering
\caption{\textbf{Core-protocol baselines grouped by structural input.}
SO-OPF and bilinear factorial rows receive the axis taxonomy; learned
$Q$ uses flat cell labels; supervised probes receive native axis
labels.  Pixel and random-init carriers use factorial assignment and
are reported only on rebuilt COCO.  Entries are INJ means over ten
seeds; variability is omitted here and reported for the assignment
comparison in Table~\ref{tab:headline}.}
\label{tab:baselines}
\small
\setlength{\tabcolsep}{3.6pt}
\begin{tabular}{@{}lccc@{}}
\toprule
\rowcolor{black!6}
Method & Shapes3D & MuJoCo & COCO\\
\midrule
\multicolumn{4}{@{}l}{\textit{Known factorial assignment}}\\
SO-OPF                 & .874 & .452 & .799\\
Bilinear, factorial $Q$ & .742 & .428 & .601\\
\midrule
\multicolumn{4}{@{}l}{\textit{Flat observed-cell labels}}\\
SO-OPF, learned $Q$   & .769 & .569 & .762\\
\midrule
\multicolumn{4}{@{}l}{\textit{Native-axis supervised probes}}\\
MLP probe              & .056 & .004 & .049\\
Attention probe        & .160 & .024 & .117\\
Oracle-region operation & .784 & .387 & .310\\
\midrule
\multicolumn{4}{@{}l}{\textit{Alternative COCO carriers, factorial $Q$}}\\
Random-init ViT        & -- & -- & .573\\
Pixels                  & -- & -- & .503\\
\bottomrule
\end{tabular}
\end{table}

Unstructured probes do not solve the held-out binding task.  On COCO,
frozen DINOv3 with factorial assignment exceeds both the random-init
and pixel carriers, while those two carriers remain above the
supervised probes.  The result therefore depends on both
representation and readout structure.

\paragraph{Factored versus dense carrier.}
The bilinear control trails SO-OPF on all three datasets under
factorial assignment.

\begin{table}[!htbp]
\centering
\caption{\textbf{Shapes3D carrier comparison (means over $n=10$ seeds).}
Core rows use Eq.~\ref{eq:core_objective}; enhanced rows use
Eq.~\ref{eq:enh_objective}.  The enhanced op-inj term consumes native
operation labels, so enhanced learned-$Q$ rows are not flat-label-only
experiments.  Non-inj and INJ are many-to-one and injective scores;
Coll. uses threshold 0.5.  Variability is omitted from this compact
carrier comparison.}
\label{tab:carrier}
\small
\setlength{\tabcolsep}{4.0pt}
\begin{tabular}{@{}llllcc@{}}
\toprule
\rowcolor{black!6}
Carrier & Assignment & Protocol & Non-inj & INJ & Coll.\\
\midrule
Factored & factorial & core & .874 & .874 & .022\\
Factored & learned & core & .780 & .769 & .156\\
Bilinear & factorial & core & .742 & .742 & .100\\
\midrule
Factored & factorial & enhanced & .913 & .913 & .000\\
Factored & learned & enhanced & .841 & .841 & .000\\
Dense & learned & enhanced & .724 & .653 & .267\\
\bottomrule
\end{tabular}
\end{table}

Under matched enhanced supervision
(Table~\ref{tab:carrier}), factorization removes the dense carrier's
0.071 laundering gap and improves learned-$Q$ INJ from 0.653
to 0.841.  With factorial assignment, the enhanced factored carrier
reaches 0.913; the paired factorial-minus-learned difference is 0.072
($p=2.0\times10^{-8}$).  These results support the carrier design and
the value of known structure.  They do not support flat-label
discovery because the enhanced objective uses native operation labels.

\paragraph{Additional fixed-structure tests.}
Under known factorial assignment, SO-OPF reaches $0.870\pm0.013$ on a
$3\times5$ grid and $0.843\pm0.006$ joint accuracy for two simultaneous
edits.  Full protocols and diagnostics are reported in the
supplementary material.

\section{Limitations}
\label{sec:limitations}

\paragraph{What is recovered.}
The core learned-$Q$ arm studies weakly supervised assignment recovery:
grid dimensions, flat cell labels, and the intervention construction
are known, but separate axis labels are not.  Factorial assignment
receives the full support--operation taxonomy, and the enhanced op-inj
objective receives native operation labels; those settings answer
different questions.  Learned assignment retains much of the
factorial-$Q$ score on Shapes3D and COCO, including the SigLIP2
replication on COCO, but its lower mean, greater across-seed variation,
and imperfect assignment recovery preclude an identifiability claim.
No experiment here constitutes unsupervised representation learning or
proves equivalence to known structure.

\paragraph{Where the boundary lies.}
On MuJoCo, learned assignment scores 0.569 with DINOv3 and 0.484 with
SigLIP2; collapse remains 43.3\% and 51.1\%, respectively, while
factorial assignment also performs poorly.  Two encoders make this a
more credible boundary, but do not isolate whether it comes from the
intervention geometry (material-space hue and inversion producing
similar token signatures), the frozen representation, or their
interaction.  More broadly, the tested operations are low-level
appearance edits and the supports are predefined categories.  Semantic
actions, moving or unstructured supports, and end-to-end encoder
adaptation remain untested.

\section{Conclusion}
\label{sec:conclusion}

Compositional readout depends separately on the carrier and on the
assignment from cell labels to a factorial grid.  SO-OPF reduces
operation laundering relative to a dense carrier and, with known
factorial assignment, composes held-out cells on Shapes3D and COCO.
Learning the assignment from flat labels recovers much of this
structure, but with lower and less stable performance.  The COCO result
replicates with SigLIP2, whereas MuJoCo remains difficult for both
encoders.

Injective evaluation distinguishes composition from slot reuse.
SO-OPF is therefore a readout for controlled grids, not an
identifiability claim.

\clearpage
\appendix
\setcounter{secnumdepth}{1}
\section{Additional Protocol and Results}

\paragraph{Protocol details.}
Compressed identity does not reindex the held-out cell to the ninth
factorial coordinate.  Within each observed cell a seeded permutation
first allocates 80\% of pairs to a development pool, from which 20\%
is withheld for validation---yielding 224/56/70
optimization/validation/unused pairs per 350-pair cell (192/48/60 per
300-pair COCO~\citep{lin2014coco} cell).  For every mode, restart
selection uses flat observed-cell validation accuracy; Hungarian
alignment~\citep{kuhn1955hungarian} and held-out pairs are unavailable
during selection.

\paragraph{Core and enhanced supervision.}
The core objective uses unit-weight cell cross-entropy and
support-use entropy; localization, assignment-balance, axis-balance,
and operation-injectivity weights are zero.  The enhanced objective
adds localization (weight 1), assignment balance (0.1), axis balance
(0.5), and operation injectivity (1), with localization margin
$\rho=0.6$.  Operation injectivity uses native operation labels,
making the enhanced protocol axis-aware.

\FloatBarrier
\section{Cross-Encoder Consolidated Summary}

Table~\ref{tab:cross_encoder} consolidates results on
Shapes3D~\citep{kim2019shapes3d}, COCO, and
MuJoCo~\citep{todorov2012mujoco} using
DINOv3~\citep{oquab2025dinov3} and
SigLIP2~\citep{tschannen2025siglip2}.

\begin{table}[t]
\centering
\caption{\textbf{Cross-encoder summary.}
INJ is injective cell accuracy; Coll. is collapse rate at threshold
0.5; Rec. is factorial assignment recovery.  All values are means over
$n=10$ seeds.}
\label{tab:cross_encoder}
\small
\setlength{\tabcolsep}{3.0pt}
\begin{tabular}{@{}llcccc@{}}
\toprule
\rowcolor{black!6}
Dataset & Encoder & Assign. & INJ & Coll. & Rec.\\
\midrule
Shapes3D & DINOv3 & factorial & .874 & .022 & --\\
Shapes3D & DINOv3 & learned & .769 & .156 & .900\\
Shapes3D & DINOv3 & compressed & .545 & .444 & --\\
\midrule
COCO & DINOv3 & factorial & .799 & .033 & --\\
COCO & DINOv3 & learned & .762 & .111 & .961\\
COCO & SigLIP2 & factorial & .826 & .011 & --\\
COCO & SigLIP2 & learned & .751 & .078 & .965\\
\midrule
MuJoCo & DINOv3 & factorial & .452 & .667 & --\\
MuJoCo & DINOv3 & learned & .569 & .433 & .643\\
MuJoCo & SigLIP2 & factorial & .476 & .656 & --\\
MuJoCo & SigLIP2 & learned & .484 & .511 & .676\\
\bottomrule
\end{tabular}
\end{table}

On Shapes3D and COCO, learned assignment approaches factorial
performance with low collapse and high recovery.  On MuJoCo, both
encoders show high collapse and low INJ regardless of assignment mode,
confirming that this boundary is not encoder-specific.

\subsection{Scaling and Carrier Controls}

\paragraph{Larger operation set.}
Extending Shapes3D to a $3\times5$ grid (hue, invert, pattern,
brightness, blur; 15 cells, chance $1/15$) gives INJ
$0.870\pm0.013$, zero collapse, and distinct-support rate 1.0 under
known factorial assignment.  The minimum per-cell mean is 0.586.

\paragraph{Two-support simultaneous edits.}
When two observed supports are edited simultaneously, a top-2 readout
over the energy tensor reaches joint accuracy
$0.843\pm0.006$ over five seeds (uniform chance $1/36$), with
at-least-one-correct accuracy 0.999 and support-pair agreement 0.833.
This tests superposition at inference time rather than discovery of an
unseen grid cell.

\paragraph{COCO carriers.}
On the rebuilt, globally image-disjoint COCO data, factorial SO-OPF
reaches $0.799\pm0.015$, compared with
$0.601\pm0.017$ for the bilinear carrier
\citep{tenenbaum2000separating},
$0.573\pm0.020$ for a frozen randomly initialized ViT, and
$0.503\pm0.016$ for pixels under injective alignment.  Supervised MLP,
attention, and oracle-operation probes reach 0.049, 0.117, and 0.310.

\section{Sensitivity and Failure Analysis}
\label{sec:supp_threshold}

\paragraph{Collapse threshold robustness.}
At thresholds 0.4/0.5/0.6, collapse rates for factorial versus learned
assignment are 1.1/2.2/3.3\% versus 13.3/15.6/18.9\% on Shapes3D,
0.0/3.3/18.9\% versus 2.2/11.1/23.3\% on COCO, and
66.7/66.7/66.7\% versus 42.2/43.3/43.3\% on rebuilt MuJoCo.
The method ordering and the MuJoCo boundary do not depend on choosing
exactly 0.5.

For the SigLIP2 replication, the corresponding factorial/learned rates
at thresholds 0.4/0.5/0.6 are 0.0/1.1/7.8\% versus
6.7/7.8/20.0\% on COCO and 51.1/65.6/66.7\% versus
51.1/51.1/52.2\% on MuJoCo.  COCO remains stable; MuJoCo remains
failure-prone throughout the threshold range.

\paragraph{MuJoCo per-cell breakdown.}
Hue and inversion are applied by changing renderer materials before
rerendering; pattern is a masked pixel-space overlay.  Under DINOv3
factorial assignment, pattern cells score 0.992--1.000, whereas hue
cells score 0.090--0.275 and inversion cells are similarly low.
SigLIP2 shows the same split: all three pattern cells score 1.000
while hue scores 0.236--0.439 and inversion 0.063--0.072.  The
boundary is therefore operation-specific, not uniform across cells.
This pattern is consistent with material-space operations producing
less discriminable token-level signatures than pixel-space overlays,
though the present evidence does not isolate this as the sole cause.

\paragraph{Learned assignment on MuJoCo.}
Learned assignment improves aggregate DINOv3 INJ from 0.452 to 0.569
but recovers only 64.3\% of factorial assignments and retains 43.3\%
collapse.  Under SigLIP2, the learned score drops from 0.629
(many-to-one) to 0.484 (injective), with distinct-support rate 0.856
and 51.1\% collapse.  The injective drop confirms slot reuse rather
than genuine compositional routing.

\paragraph{Statistical comparisons.}
On Shapes3D under the core protocol, factorial-minus-learned raw
accuracy is 0.095 across paired seed means
($t=4.01$, two-sided $p=0.0031$, $n=10$).  Under the enhanced
axis-aware protocol the difference is 0.072
($p=2.0\times10^{-8}$).  These tests compare assignments within a
protocol; they do not convert the enhanced learned-$Q$ result into a
flat-label-only experiment.

\section{Future Directions}

The MuJoCo boundary---where material-space operations produce less
discriminable token signatures---motivates tests with encoders trained
using renderer-diverse augmentation and with semantic operations such
as object removal or relighting.  These experiments could help
distinguish representational limitations from structural ones.
Replacing predefined supports with discovered spatial decompositions
would further move the evaluation from supervised structure recovery
toward unsupervised compositional binding.

\section{Reproducibility Details}

The main experiments use seeds 0--9, 2,000 optimization
steps, five restarts, and test-blind validation selection.  The
DINOv3 and SigLIP2 checkpoints are
\path{dinov3-vitl16-pretrain-lvd1689m} and
\path{siglip2-large-patch16-256}.  Their respective runs used NVIDIA
H200 and H100 GPUs.
The result artifacts retain assignment mode, validation selector, loss
weights, seeds, restarts, and cache provenance.  The software
environment lock, renderer configuration, preprocessing code, and
per-fit outputs will be released with the final publication.

\clearpage
{\small
\bibliography{references}
}

\end{document}